\documentclass[sigconf]{acmart}

\def\BibTeX{{\rm B\kern-.05em{\sc i\kern-.025em b}\kern-.08emT\kern-.1667em\lower.7ex\hbox{E}\kern-.125emX}}

\copyrightyear{2020}
\acmYear{2020}
\setcopyright{acmlicensed}
\acmConference[DSHealth '20]{DSHealth '20: KDD Workshop on Applied Data Science for Healthcare}{August 24, 2020}{San Diego, CA}

\usepackage{multirow}
\usepackage[ruled]{algorithm2e} 

\SetAlFnt{\small}
\SetAlCapFnt{\small}
\SetAlCapNameFnt{\small}
\SetAlCapHSkip{0pt}
\IncMargin{-\parindent}

\begin{document}

\setlength{\abovedisplayskip}{1pt}
\setlength{\belowdisplayskip}{1pt}

\title[Transfer Learning for Activity Recognition in Mobile Health]{Transfer Learning for Activity Recognition in Mobile Health}

\author{Yuchao Ma}
\authornote{Work completed while at Washington State University.}
\affiliation{%
  \institution{Amazon.com}
  \city{Seattle}
  \state{WA}
 }

\author{Andrew T. Campbell}
\affiliation{%
  \institution{Dartmouth College}
  \city{Hanover}
  \state{NH}
}

\author{Diane J. Cook}
\affiliation{%
  \institution{Washington State University}
  \city{Pullman}
  \state{WA}
}

\author{John Lach}
\affiliation{%
  \institution{George Washington University}
  \city{Washington}
  \state{DC}
}

\author{Shwetak N. Patel}
\affiliation{%
  \institution{University of Washington}
  \city{Seattle}
  \state{WA}
}

\author{Thomas Ploetz}
\affiliation{%
  \institution{Georgia Institute of Technology}
  \city{Atlanta}
  \state{GA}
}

\author{Majid Sarrafzadeh}
\affiliation{%
  \institution{University of California Los Angeles}
  \city{Los Angeles}
  \state{CA}
}

\author{Donna Spruijt-Metz}
\affiliation{%
  \institution{University of Southern California}
  \city{Los Angeles}
  \state{CA}
}

\author{Hassan Ghasemzadeh}
\authornote{Corresponding author (hassan.ghasemzadeh@wsu.edu).}
\affiliation{%
  \institution{Washington State University}
  \city{Pullman}
  \state{WA}
}
\renewcommand{\shortauthors}{Y. Ma, A. T. Campbell, D. J. Cook, J. Lach, S. N. Patel, T. Ploetz,  M. Sarrafzadeh, D. Spruijt-Metz, and H. Ghasemzadeh}

\begin{abstract}
While activity recognition from inertial sensors holds potential for mobile health, differences in sensing platforms and user movement patterns cause performance degradation. Aiming to address these challenges, we propose a transfer learning framework, \textit{TransFall}\footnote{TransFall code and sample data are available at \url{https://github.com/y-max/TransFall.}}, for sensor-based activity recognition. TransFall's design contains a two-tier data transformation, a label estimation layer, and a model generation layer to recognize activities for the new scenario. We validate TransFall analytically and empirically.
\end{abstract}
%
%



\maketitle

\section{Introduction}
The rapid integration of sensors technologies, fueled by computational algorithms, creates a unique opportunity for remote health monitoring, long-term fitness tracking, and fall detection. A core task to support these applications is activity recognition. However, differences in sensing platforms and user behavior has limited generalizability of the activity recognition models. For example, a user may replace an old mobile device with a new model. While the user wants to maintain the usability of a well-trained motion analysis app installed on the old device, the user would like to avoid providing additional manual annotations for model re-training on the new device.

\begin{figure}[tbh!]
\vspace{-3mm}
\centering
\includegraphics[width=2.6in]{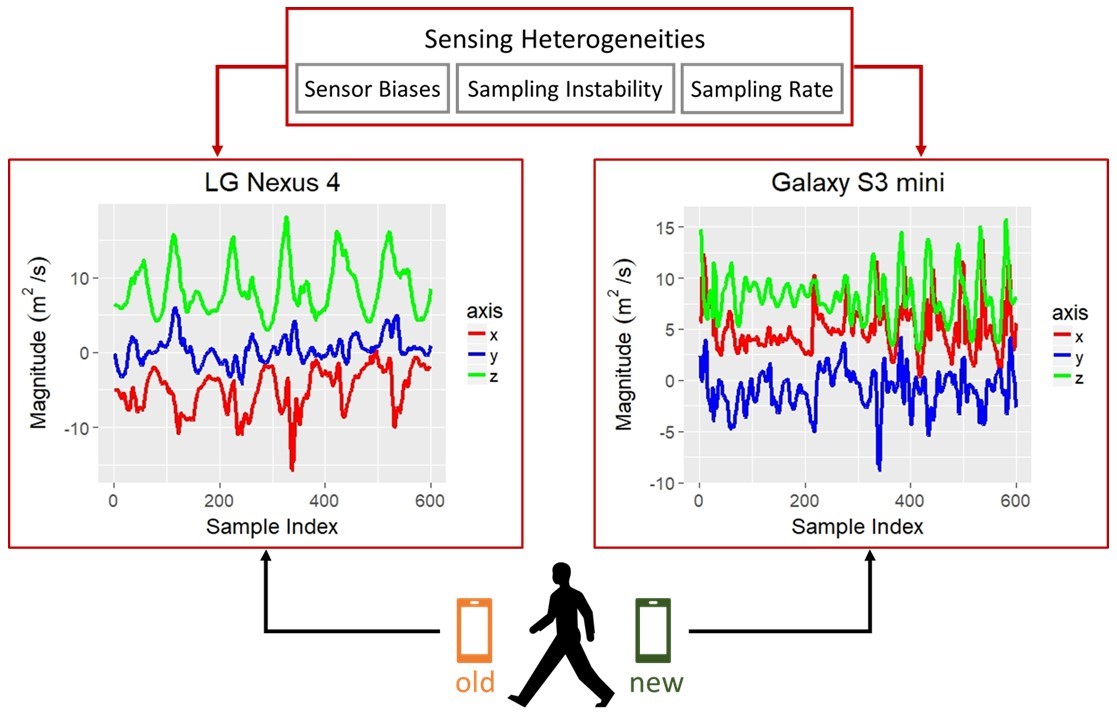}
\vspace{-2mm}
\caption{Sensor readings from two smartphone models.}
\label{fig:cross}
\vspace{-3mm}
\end{figure}

Activity recognition performance is also adversely impacted by sensor biases from low-quality modules and sampling rate instability \cite{hharshort}. For example, F1-score declines $34.4$\% when training and test data belong to different devices (e.g., Samsung Galaxy S3 vs LG Nexus). Figure \ref{fig:cross} shows acceleration for one subject's walking behavior gathered by two smartphones. Such data divergence can propagate through the data processing pipeline, leading to accuracy decline.

\begin{figure}[tbh!]
\centering
\includegraphics[width=2.6in]{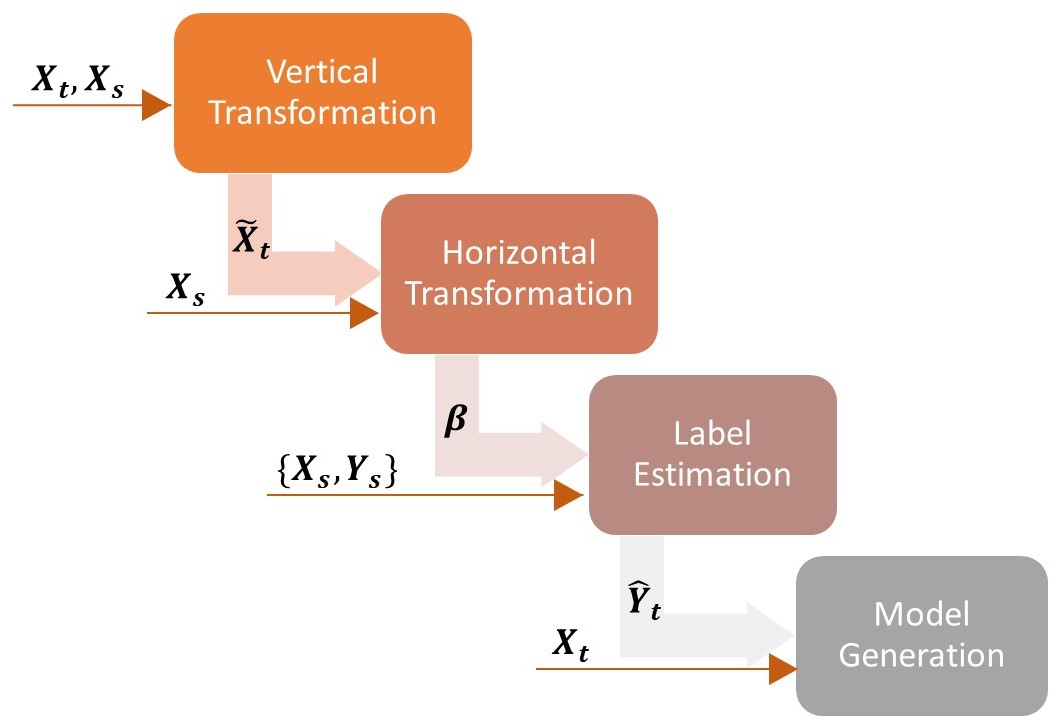}
\vspace{-2mm}
\caption{TransFall's sequential transfer learning design.}
\label{fig:frame}
\vspace{-3mm}
\end{figure}

We propose \textit{TransFall} to overcome the challenges caused by cross-domain variations while reducing the dependency on labeled training data. Shown in Figure \ref{fig:frame}, the framework starts with a two-tier data transformation layer based on marginal distribution matching approaches, followed by a label estimation layer using a kernel method encoded in a weighted least-mean-squares fitting, and ends with a model generation layer using the previous step's label set.

\begin{figure}[tbh!]
\centering
\includegraphics[width=\linewidth]{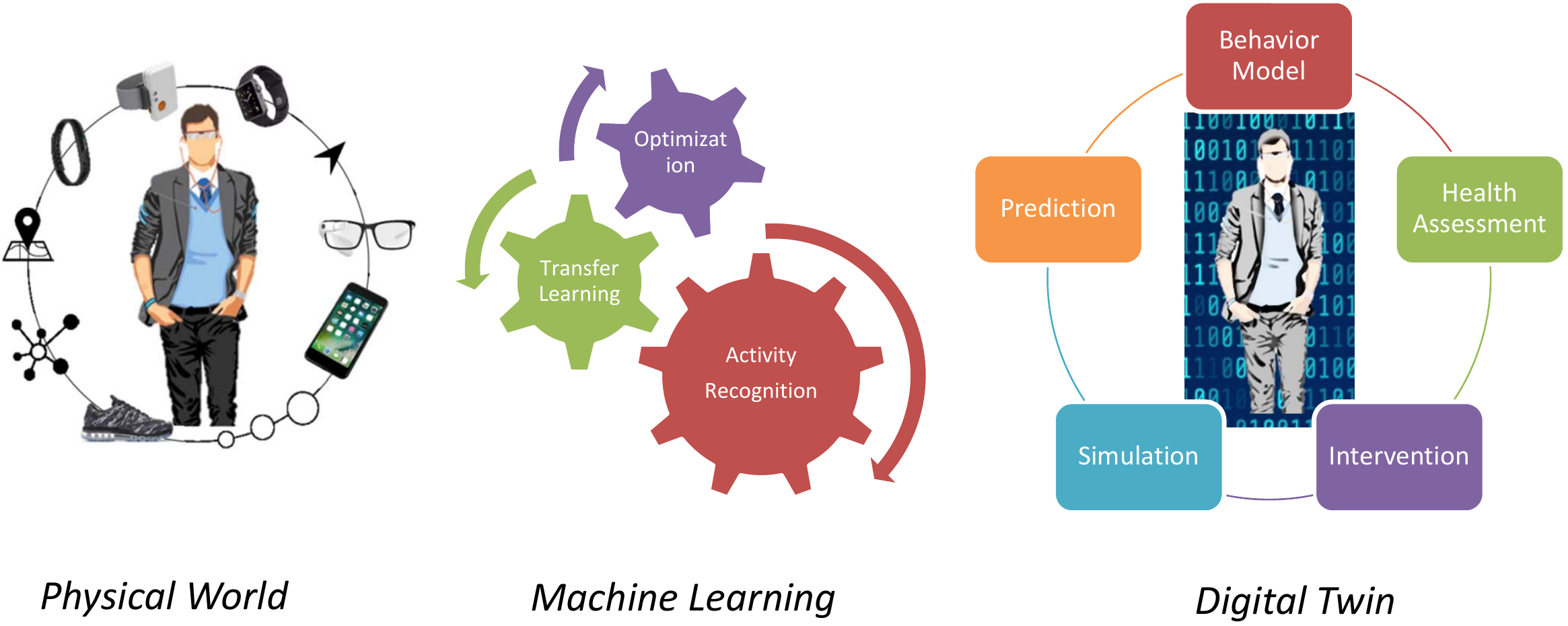}
\vspace{-2mm}
\caption{Digital twin vision.}
\label{fig:dt}
\vspace{-3mm}
\end{figure}

Development of transfer learning algorithms for activity recognition is also central in the long-term goal of creating a person's {\em digital twin}. A digital twin is a digital replica of the human subject, built from multiple information sources including mobile sensor data. This quantified self provides a platform to understand the relationship between human behavior and influencing factors including health, genetics, and the environment. A digital twin offers the basis for automating health assessment and evaluating potential interventions on digital prototypes. Fusing data from multiple information sources, and building a model that is robust enough to accurately depict a diverse population, relies on the ability to effectively transfer data between domains. TransFall thus represents one component for completing the digital twin vision.

\section{The Transfer Learning Framework}
\subsection{Problem Statement}\label{sec:problem}
Let $X_s$ be a set of $N_s$ labeled data samples from source domain $\mathcal{S}$, where $x^s \in X_s$ is a $d$-dimensional variable drawn from marginal distribution $P_s(x)$ and label set $Y_s$ represents $L$ activities. Furthermore, let $X_t$ be a set of $N_t$ unlabeled data samples from target domain $\mathcal{T}$, and $x^t \in X_t$ be drawn from marginal distribution $P_t(x)$ with the same set of activities. TransFall offers an activity recognition model $\mathcal{M}: \mathcal{X} \mapsto \mathcal{Y}$ capable of accurately estimating the corresponding labels for the data samples collected on $\mathcal{T}$. With changes to the sensing platform, there exists a distribution shift in covariates $x$ between $X_s$ and $X_t$, resulting in $P_s(x) \neq P_t(x)$. Therefore, the first task is to address the covariate shift through data transformation.

\subsection{Vertical Transformation}\label{sec:vertical}
The vertical transformation layer matches the marginal distributions of individual column variables between $X_s$ and $X_t$. Due to the multi-dimensional nature of data sample $x \in \mathcal{R}^d$, the marginal distribution $P(x)$ can be determined by a joint probability distribution $P(x_1, \cdots, x_d)$. Raw signals are first converted into vector objects by projecting the input signals onto a designated feature space $\mathcal{F}$. Each feature $f_i \in \mathcal{F}$ is computed independently. Hence, we use the naive Bayes approximation to factorize the joint probability distribution, or
$P(x) = P(x_1, \cdots, x_d) = \prod_{i = 1}^{d} P(x_i)$.

Therefore, the optimization objective of TransFall can be viewed as the summation of $d$ subordinate optimization problems for $d$ column variables, where $\phi_i \in \Phi$ is a mapping function that converts the original probability distribution of a one-dimensional variable into a different distribution.

\begin{equation}
\underset{\Phi}{\text{Minimize}} \sum_{i = 1}^{d} \int_{x_i} |P_s(x_i) - \phi_i (P_t(x_i))| dx_i
\end{equation}

In practice, the true distribution of a random variable is unattainable, and hence the normal distribution is commonly adopted to approximate the marginal distribution of sensor data. As a result, each column variable $x_i$ is assumed to be drawn from a normal distribution with mean $\mu_i$ and variance $\sigma_i^2$, denoted as $x_i \sim \mathcal{N}(\mu_i, \sigma_i^2)$.

To minimize the difference of the means and variances between column variables in $X_s$ and $X_t$, we perform a linear transformation on each dimension $i \in [1,d]$ in $X_t$ with respect to $X_s$. The output of the vertical transformation is the transformed target dataset $\tilde{X_t}$, where $\tilde{x}_i^t \sim \mathcal{N}(\mu_i^s, {\sigma_i^s}^2)$ for each $i \in [1,d]$. 

\subsection{Horizontal Transformation} \label{sec:horizontal}
The horizontal transformation layer further reduces the discrepancies in multi-variate variables between $X_s$ and $\tilde{X_t}$ using importance sampling, a common method to address covariate shift \cite{imsample-1}. This technique finds a weight factor $\beta$ for $X_s$ that assigns higher weight to those source data samples that are more representative of the target dataset. The goal of this transformation is as follows:

\begin{equation}\label{eq:opt2}
\underset{\beta}{\text{Minimize}} \int_x |\beta(x)P_s(x) - P_t(x)| dx \; .
\end{equation}

Because distributions $P_s(x)$ and $P_t(x)$ are unknown in practice, we use a kernel-based algorithm, empirical Kernel Mean Matching (eKMM) \cite{datashift}, to find the optimal weight factor $\beta$ with the use of Reproducing Kernel Hilbert Space (RKHS) technique.

Let $\Phi: \mathcal{X} \mapsto \mathcal{F}$ be a function that maps a vector variable $X$ onto a feature space $\mathcal{F}$. The output of the eKMM algorithm is the optimal weight factor $\beta$, which can minimize the distance between the empirical means of $X_s$ and $\tilde{X_t}$ on the feature space $\mathcal{F}$, as shown in the following equations.

\begin{equation}\label{eq:kmm}
\begin{aligned}
& \underset{\beta}{\text{Minimize}} 
& & \| \frac{1}{N_s}\sum_{i = 1}^{N_s}\beta_i \Phi(x_i^s) - \frac{1}{N_t}\sum_{j = 1}^{N_t}\Phi(x_j^t) \|^2 \\
& \text{Subject to}
& & \beta_i \geq 0 , \; i \in [1, N_t] \\
&&& |\frac{1}{N_t} \sum_{i = 1}^{N_t}\beta_i - 1| \leq \epsilon.
\end{aligned}
\end{equation}

The first constraint in (\ref{eq:kmm}) refers to the non-negative probability property and the second constraint guarantees that the re-weighted distribution $\beta(x)P_s(x)$ is close to a valid probability distribution that sums to $1$. We use the RKHS technique \cite{rkhs-2} to solve the optimization problem in (\ref{eq:kmm}) based on an important property:

\begin{proposition}
Given a positive definite kernel $k$ over a vector space $\mathcal{X}$, we can find a Hilbert space $\mathcal{H}$ and a mapping function $\Phi: \mathcal{X} \mapsto \mathcal{H}$, such that
$k(x_i, x_j) = \langle \Phi(x_i), \Phi(x_j) \rangle_{\mathcal{H}}$, where $x_i, x_j \in \mathcal{X}$.
\end{proposition}

With the use of a kernel function that is positive definite on Euclidean space $\mathcal{R}^{d}$, optimization (\ref{eq:kmm}) can be solved without explicitly defining the mapping function $\Phi$. For this purpose, we use Gaussian kernel. Therefore, the objective function (\ref{eq:kmm}) can be rephrased as:

\begin{equation}\label{eq:quad}
\underset{\beta}{\text{Minimize }} \frac{1}{2} \beta^\top K \beta - \kappa^\intercal \beta
\end{equation}
\noindent where the kernel matrix $K$ and the kernel expansion $\kappa$ are given by
\begin{equation} \label{eq:kernelmatrix}
\begin{aligned}
K_{ij} & := k(x_i^s, x_j^s); \;\;\;\;\;\;
\kappa_i & := \frac{N_s}{N_t} \sum_{j = 1}^{N_t} k(x_i^s, x_j^t)
\end{aligned} 
\end{equation}

As a result, the optimal $\beta$ can be determined by solving (\ref{eq:quad}) with the constraints listed in (\ref{eq:kmm}) using quadratic programming. Note that, similar to the vertical transformation module, the horizontal transformation module has also the potential to be coupled with existing machine learning algorithms that support the sample re-weighting.

\subsection{Label Estimation}\label{sec:label}
Given the transformed target dataset $\tilde{X_t}$ and the weight factor $\beta$, which approximates the distribution of $\tilde{X_t}$ using the source dataset $X_s$, the label estimation module intends to estimate the label set $\hat{Y}_t$ for $\tilde{X_t}$ in preparation for training an activity recognition model for the target domain. The label estimation objective can be written as follows, for $x_i \in X_s$: \[\underset{f}{\text{Minimize}} \sum_{i = 1}^{N_s} |y_i - \beta(x_i)f(x_i)|\]

We can rewrite this optimization problem using a weighted least-mean-squares (LMS) fitting technique with a 2-norm regularization term as shown below. The LMS technique is commonly used for parameter estimation in linear models \cite{datashift}.

\begin{equation}\label{eq:minf}
\underset{f}{\text{Minimize }} \sum_{i = 1}^{N_s} \beta_i (y_i^s - f(x_i^s))^2 + \lambda \|f\|^2
\end{equation}

However, the optimal function $f$ in (\ref{eq:minf}) is not necessarily linear. Therefore, we convert (\ref{eq:minf}) using a linear model, based on the representer theorem \cite{representer}, to the following:
\begin{displaymath}
\underset{\alpha}{\text{Minimize }} \sum_{i = 1}^{N_s} \beta_i (y_i^s - \sum_{j = 1}^{N_s}\alpha_j k(x_j^s, x_i^s))^2 + \lambda \|\sum_{j = 1}^{N_s} \alpha_j k(x_j^s, \cdot)\|^2
\end{displaymath}
which can be written (after extension) as shown in (\ref{eq:mina}).
\begin{equation}\label{eq:mina}
\underset{\alpha}{\text{Minimize }} (Y_s - K\alpha)^\top \overline{\beta} (Y_s - K\alpha) + \lambda \alpha^\top K\alpha
\end{equation}
\noindent where $K$ represents the kernel matrix in (\ref{eq:kernelmatrix}) and $\overline{\beta}$ is a $N_s \times N_s$ diagonal matrix of $\beta$. If $K$ and $\overline{\beta}$ are full rank matrices, the optimal solution for $\alpha$ can be derived using:
\begin{equation}\label{eq:aresult}
\alpha = (\lambda {\overline{\beta}}^{ -1} + K)^{-1}Y_s 
\end{equation}

Because the label set in activity recognition often contains multiple activity classes, we use a one-to-all approach for label estimation, by first solving $L$ optimal linear models $\alpha^m$ for all activity labels, and then combining $L$ corresponding estimations of the data sample $x_i^t \in \tilde{X_t}$, to make the final prediction.
\begin{equation}
\hat{y}^t_i = \underset{m}{\text{argmax }} \sum_{j = 1}^{N_s} \alpha_j^m k(x_j^s, x_i^t)
\end{equation} 

\begin{table}[t!]
\caption{Dataset and experiment abbreviations.}
\small
\label{tab:notation}
\centering
\begin{tabular}{c|c|l}
\hline
&\textbf{Notation} & \multicolumn{1}{c}{\textbf{Description}} \\
\hline
\multirow{3}{*}{Dataset} & Phone & 8 smartphones and 9 subjects\\
& Watch &  4 smartwatches and 5 subjects\\
& HART & one smartphone and 30 subjects\\
\hline
\multirow{4}{*}{Cross-Platform} & P2P-S & Same-model phone-to-phone\\
& P2P-D & Different-model phone-to-phone\\
& W2W-S & Same-model watch-to-watch\\
& W2W-D & Different-model watch-to-watch\\
\hline
\multirow{6}{*}{Comparison Group} & NN & Nearest neighbor\\
& DT & Decision tree\\
& LR & Logistic regression\\
& SVM & Support vector machine\\
& Upper & Using ground truth data\\
& IWLSPC & Method in \cite{tl-23} \\
\hline
\end{tabular}
\end{table}

\section{Experimental Results}\label{sec:results}
We conducted experiments on three publicly-available datasets \cite{hharshort, hart2short}, as listed in Table \ref{tab:notation}. We evaluated the performance of TransFall in three scenarios including \textit{cross-platform}, \textit{cross-subject}, and \textit{hybrid}. We use the notations shown in Table \ref{tab:notation} to refer to various datasets, transfer learning scenarios, and comparison approaches. The comparison approach IWLSPC refers to the importance-weighted least-squares probabilistic classifier (IWLSPC) approach introduced in \cite{tl-23}. IWLSPC combines a least-squares probabilistic model with a sample re-weighting approach, to handle the changes in data distribution between the source and the target datasets. While the sample re-weighting technique in TransFall is similar to that of IWLSPC, TransFall performs a two-tier data transformation on both datasets, to empirically match the distributions of the two datasets for more accurate label estimation.

\begin{figure}[tbh!]
\centering
\includegraphics[width=\linewidth]{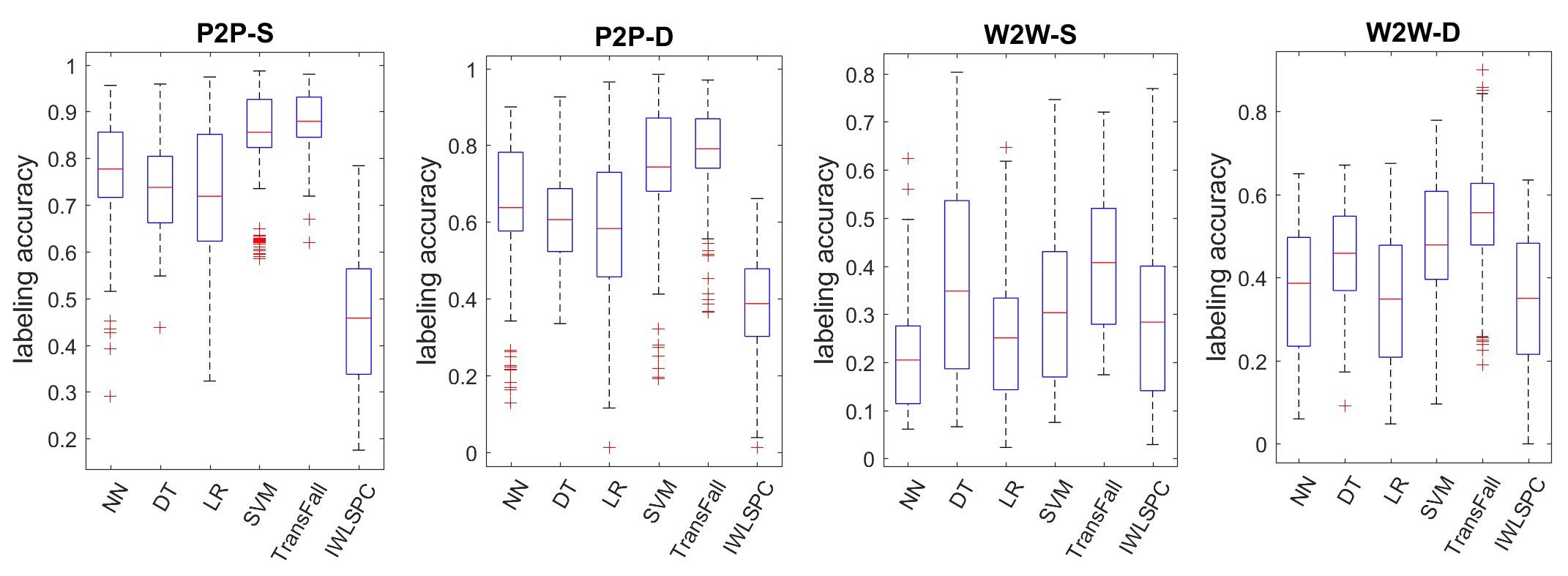}
\vspace{-2mm}
\caption{Results for cross-platform scenario sub-cases.}
\label{fig:label-cp}
\vspace{-5mm}
\end{figure}

\begin{figure}[tbh!]
\centering
\includegraphics[width=\linewidth]{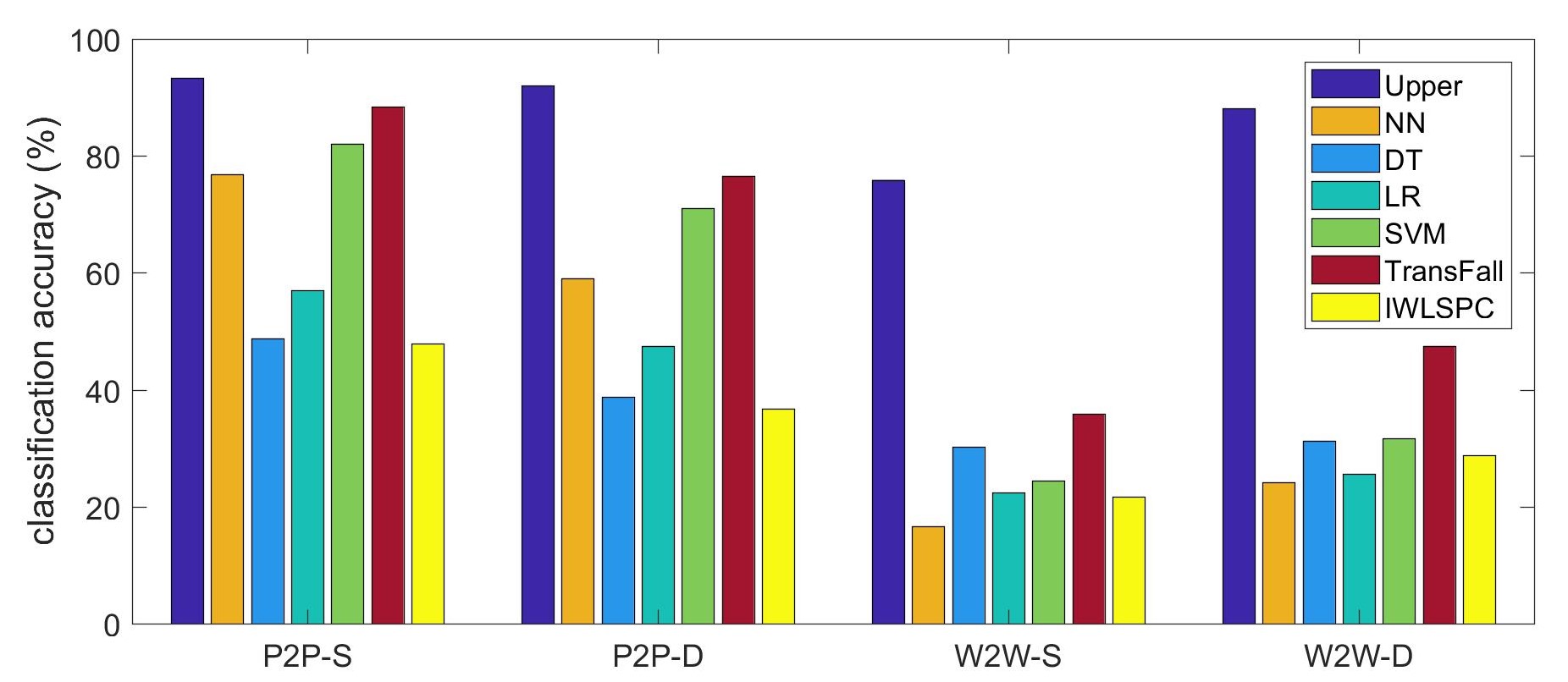}
\vspace{-2mm}
\caption{Results for cross-platform scenario sub-cases.}
\label{fig:ar-cp}
\vspace{-5mm}
\end{figure}

\subsection{Cross-Platform Transfer Learning}
\label{sec:crossp}
This scenario refers to the case when source data and target data are gathered using different devices. We further divide this scenario into four cases as shown in Table \ref{tab:notation}.

Figure \ref{fig:label-cp} shows the results of label estimation using different approaches in the four sub-cases accordingly, where the red central mark on each box indicates the mean value of labeling accuracy, the bottom and top edges indicate the 25\textsuperscript{th} and 75\textsuperscript{th} percentiles respectively, and the outliers are denoted as plus symbols.

Overall, all the approaches achieve a higher accuracy on the data gathered by smartphones than that of smartwatches, mainly due to the less stability of data collection using smartwatches. Given the data gathered by smartphones, the labeling accuracy is higher in P2P-S case than that of P2P-D case. This result is coincident with general expectation, because source device and target device are in different models for P2P-D case, which results in large diversity between the obtained datasets due to the differences, such as sampling frequency and platform configuration.

Compared to the other five approaches shown in Figure \ref{fig:label-cp}, TransFall achieves the highest labeling accuracy in average, with a correct labeling rate of $0.88$, $0.79$, $0.41$ and $0.56$ in the four cases respectively. The increase in the labeling accuracy of TransFall compared to other approaches is $>2.7$\% for P2P-S, $>6.3$\% for P2P-D, $>16.6$\% for W2W-S, and $>16.1$\% for W2W-D.

In this task, logistic regression model was empirically chosen to carry out activity recognition after label information transfer. Figure \ref{fig:ar-cp} shows the results in four transfer learning cases. In general, the classification accuracy is consistent with the labeling accuracy, because the quality of training dataset is determined by the precision of labeling the target data in previous task.

In Figure \ref{fig:ar-cp}, the performance upper bound (referred to as ``Upper'') of machine learning model trained with the ground truth appears to be dramatically lower on the data gathered by smartwatches than that of smartphones. The accuracy upper bound is $93.3$\% in P2P-S case, but $75.9$\% in W2W-S case. This performance decline reflects the lower quality of sensor data samples collected using smartwatches. 

TransFall still achieves improved performance over other approaches. Its classification accuracy is $88.4$\%, $76.6$\%, $35.9$\% and $47.5$\% for the four sub-cases, respectively. Moreover, the performance improvement of the machine learning model trained by TransFall is $>7.7$\% compared to the machine learning models trained by other approaches on smartphone data, and $>19$\% on smartwatch data.

\begin{figure}[tbh!]
\centering
\includegraphics[width=\linewidth]{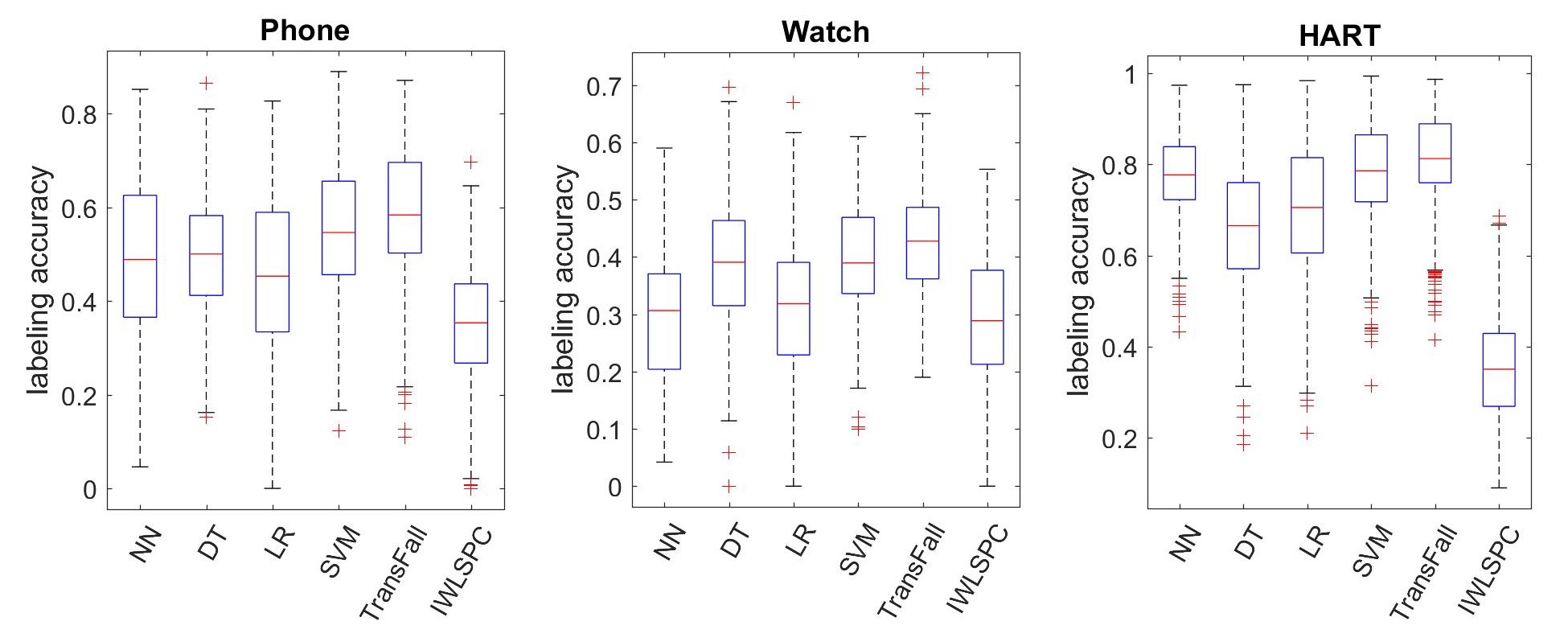}
\vspace{-2mm}
\caption{Results of cross-subject activity recognition.}
\label{fig:label-cs}
\vspace{-4mm}
\end{figure}

\begin{figure}[tbh!]
\centering
\includegraphics[width = \linewidth]{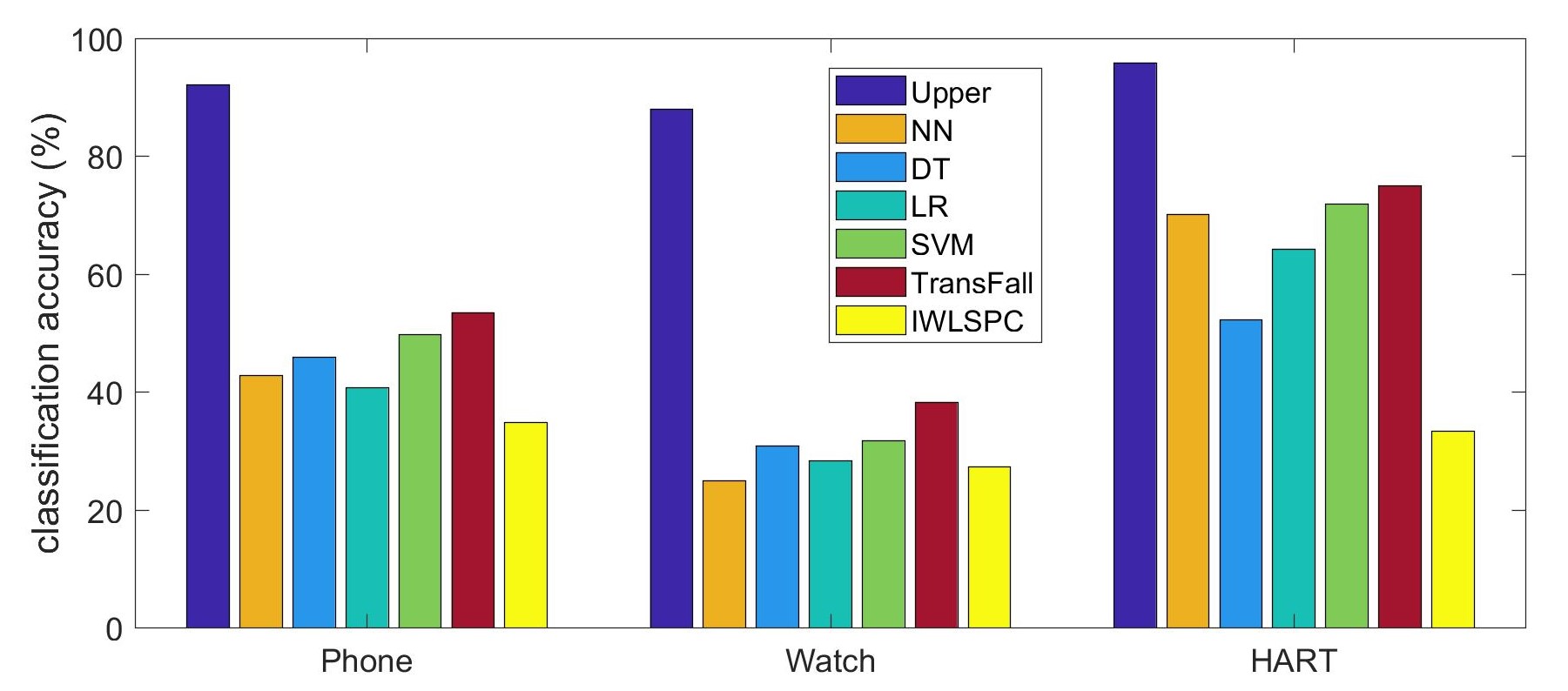}
\vspace{-2mm}
\caption{Results of cross-subject activity recognition.}
\label{fig:ar-cs}
\vspace{-4mm}
\end{figure}

\subsection{Cross-Subject Transfer Learning}
\label{sec:crosss}
In this scenario, source data and target data are collected from two subjects using the same type of mobile device. Figure \ref{fig:label-cs} shows the results of label estimation on each dataset separately. Almost all the approaches perform best on the HART dataset than the other two datasets. One possible explanation is less noise in the HART dataset which results in a more informative classification task. Moreover, TransFall performs better than other approaches in the comparison group, with an accuracy increase of $>6.9$\% on the Phone dataset, $>9.5$\% on the Watch dataset, and $>3.4$\% on the HART dataset.

Given label sets obtained using the alternative approaches, we can examine the performance of activity models trained with each label set. Figure \ref{fig:ar-cs} shows the activity recognition results. Again, the most approaches perform best on the HART dataset. The empirical upper bound of classification accuracy is $95.9$\% on the HART dataset, $92.1$\% on the Phone dataset and $88.0$\% on the Watch dataset. TransFall still exhibits the best recognition performance, with an accuracy increase of $>7.4$\% on Phone dataset, $>20.2$\% on the Watch dataset, and $>4.4$\% on the HART dataset.

\section{Conclusion and Future Work}
\label{sec:conclusion}
TransFall integrates a two-tier data transformation to perform transfer learning. Experimental results demonstrate that TransFall can steadily improve activity recognition accuracy comparing to several alternative approaches. A limitation of TransFall is the assumption of consistent feature alignments between source and target datasets. Our future work involves improving the generalizability of the current framework to handle variations in feature space between source and target datasets.

\bibliographystyle{ACM-Reference-Format}
\bibliography{kddref}
\end{document}